\documentclass{article} 
\usepackage{iclr2027_conference,times}

\usepackage{amsmath,amsfonts,bm}

\def\eqref#1{equation~\ref{#1}}

\def\1{\bm{1}}

\DeclareMathAlphabet{\mathsfit}{\encodingdefault}{\sfdefault}{m}{sl}
\SetMathAlphabet{\mathsfit}{bold}{\encodingdefault}{\sfdefault}{bx}{n}

\usepackage{hyperref}
\usepackage{url}
\usepackage{booktabs}       
\usepackage{amsfonts}       
\usepackage{nicefrac}       
\usepackage{microtype}      
\usepackage{xcolor}         

\usepackage{amsmath}        
\usepackage{amssymb}        
\usepackage{amsthm}         
\usepackage{bm}             
\usepackage{algorithm}      
\usepackage{algpseudocode}  
\usepackage{graphicx}
\usepackage{subcaption}     
\usepackage{wrapfig}
\usepackage{multirow}

\newcommand{\ganc}{g_{\mathrm{anc}}}
\newcommand{\gfr}{g_{\mathrm{fr}}}
\newcommand{\Stream}{\mathcal{D}_{\mathrm{stream}}}
\newcommand{\Real}{\mathbb{R}}
\DeclareMathOperator{\logdet}{\log\det}

\title{Fisher-Guided Submodular Data Selection for Continual Pre-Training of Large Language Models}

\author{Zhenghao Zhao\textsuperscript{1} \qquad Gaowen Liu\textsuperscript{2} \qquad Zhiling Lan\textsuperscript{1} \qquad Yan Yan\textsuperscript{1} \\
    \textsuperscript{1}University of Illinois Chicago \qquad
    \textsuperscript{2}Cisco Research\\
}

\iclrfinalcopy 
\begin{document}

\maketitle
\lhead{Preprint.}

\begin{abstract}
    Data selection is already a central bottleneck in large-language-model training, where web-scale corpora are noisy and token budgets are finite. In continual pre-training (CPT), it becomes a forgetting-control problem: a poorly chosen target-domain corpus can overwrite capabilities encoded in the pretrained checkpoint. Existing CPT practice either scores candidates with parameter-agnostic scalars such as perplexity, or mitigates forgetting by spending many extra general-domain replay tokens. Neither strategy directly asks how training on a candidate will move the model parameters.
    We show that loss-based selection causes the post-CPT Fisher diagonal to drift downward on exactly the high-Fisher coordinates the pretrained model had committed to, while leaving low-Fisher coordinates largely untouched. This asymmetry exposes a parameter-space mechanism for catastrophic forgetting. Motivated by this observation, we propose a Fisher-aware CPT selector that decomposes each candidate's gradient into an anchor component, which measures perturbation along committed parameter directions, and a frontier component, which measures update capacity in unconstrained low-Fisher subspaces. We aggregate these signals with a log-determinant submodular objective and optimize it in a single pass using a scalable streaming data selection pipeline.
    On TinyLlama-1.1B and Llama-3.1-8B CPT over medical data, our selector improves target-domain quality while bounding forgetting on held-out pretraining benchmarks. Most importantly, it is substantially more token-efficient than forgetting-aware replay. 1B selected tokens already outperform the replay strategy trained with 10B tokens on both adaptation and forgetting, giving a $10\times$ token-efficiency advantage. 
\end{abstract}    
\section{Introduction}
\label{sec:intro}

Data selection~\citep{xie2023data,xie2023doremi,zhou2023lima,chen2023maybe,chen2023alpagasus} is one of the few practical levers for making large-language-model training efficient: web-scale corpora are noisy, token budgets are finite, and the examples that enter training determine what knowledge is acquired per unit compute. The lever becomes even more consequential in continual pre-training~\citep{ibrahim2024simple,chen2025towards} (CPT), where a pretrained LLM is updated on a new domain, language, or temporal slice rather than trained from scratch \citep{ibrahim2024simple,chen2025towards}. In CPT, data selection is not only about learning faster; it is also a forgetting-control mechanism. A poorly chosen target-domain update can silently overwrite the very capabilities that made the pretrained checkpoint valuable \citep{kirkpatrick2017overcoming,li2024examining,shi2025continual}. Thus a CPT selector must solve two coupled problems at once: acquire new knowledge from the target domain while preserving the old knowledge already encoded in the model.

The standard toolkit is poorly matched to this requirement. Quality classifiers and rule-based filters \citep{peng2025dataman,calian2025datarater,marion2023less} score a candidate by cleanliness or topical fit, while per-sample loss and perplexity \citep{sorscher2022beyond} treat the model as a black box whose only output is a scalar. These signals describe a document under the current model, but they do not say how training on it would move the parameters. Replay-based CPT mitigates forgetting from another direction by mixing general-domain data back into training, but it pays for retention with many additional tokens and still leaves the target-domain selection problem unresolved. What is missing is a selector that can identify target-domain examples that are both useful for adaptation and structurally unlikely to interfere with retained capability.

Our starting point is a simple parameter-space observation. The Fisher information~\citep{kirkpatrick2017overcoming} of a pretrained model encodes, direction by direction, how strongly the predictive distribution depends on perturbing the parameters. We empirically find that parameter-agnostic method cause the post-CPT Fisher diagonal to drift downward on exactly the high-Fisher coordinates the pretrained model had committed to, while leaving low-Fisher coordinates largely untouched. This is precisely the asymmetry of catastrophic forgetting: committed directions are damaged, and uncommitted directions are underused. The observation suggests that selection should be performed not only in data space or loss space, but in the Fisher geometry of the pretrained model.

In this paper, we propose to use the Fisher information to decide which candidates enter CPT. For each candidate, we decompose its gradient into two parameter-aware components. An anchor component quantifies perturbation along high-Fisher directions that carry retained capability, while a frontier component captures update capacity in low-Fisher subspaces where the pretrained model is less committed. An effective CPT sample must balance both: it should avoid destructive interference with existing capability while supplying genuinely new, non-redundant information. This curvature-aware decomposition turns forgetting control and acquisition into a single selection signal rather than two post-hoc training heuristics.

To select a subset rather than rank candidates independently, we aggregate the anchor and frontier signals with a log-determinant objective. The resulting utility rewards gradient volume in the two Fisher-induced geometries, so it favors candidates that are high-value and mutually non-redundant. This objective is monotone submodular, so offline greedy selection attains the classical $(1 - 1/e)$ approximation guarantee of \citep{nemhauser1978analysis}. For large-scale CPT streams, we maximize the same objective in one pass with Sieve-Streaming \citep{badanidiyuru2014streaming}, restrict the decomposition to the LoRA \citep{hu2022lora} adapter subspace, replace the full Fisher with its diagonal, project gradients via TRAK \citep{park2023trak}, and use Sherman--Morrison rank-1 updates of the per-sieve Gram inverses to reduce the per-candidate cost to quadratic in a small projected dimension.

Empirically, on CPT of TinyLlama-1.1B and Llama-3.1-8B over medical data, our selector improves target-domain quality while bounding degradation on held-out pretraining benchmarks. The most direct comparison is against forgetting-aware replay: on TinyLlama-1.1B, our 1B-token selected subset already outperforms the replay strategy trained with 10B tokens on both adaptation and forgetting, giving a $\ge 10\times$ token-efficiency advantage. 

The contributions of this paper are summarized as follows:
\begin{itemize}
    \item We propose decomposing each candidate's gradient into an {anchor} component and a {frontier} component using the pretrained model's diagonal Fisher information, yielding a parameter-aware signal that simultaneously measures interference with existing knowledge and acquisition of new knowledge.
    \item We aggregate the two components into a $\logdet$ Gram-volume set function and instantiate the classical Sieve-Streaming algorithm to maximize it in a single pass over the candidate pool, with per-sample cost independent of pool size.
    \item On TinyLlama-1.1B and Llama-3 8B, our selector improves target-domain quality at a fraction of the training tokens used by random and perplexity baselines, while bounding forgetting on held-out benchmarks.
\end{itemize}

\section{Related Work}
\label{sec:related}

We give a compressed overview here and defer detailed discussion to Appendix~\ref{app:related}.

\paragraph{Continual pre-training of LLMs.}
Continual pre-training (CPT) adapts a pretrained LLM to a new domain, language, or temporal slice while combating catastrophic forgetting. One line stabilizes the optimizer through learning-rate re-warming, re-decay, and replay \citep{ibrahim2024simple,chen2025towards}; another targets domain specialization in finance \citep{xie2024efficient}, mathematics and coding \citep{fujii2025rewriting}, with scaling-law analyses given by the D-CPT law \citep{que2024d}. Forgetting during CPT~\citep{li2024examining,yildiz2024investigating,shi2025continual} of aligned/instruction-tuned models has been characterized, and mitigated via routing and model-merging \citep{chen2023lifelong,yang2024embracing,ilharco2022editing,wortsman2022robust,kleiman2025soup}. CPT has also been extended to knowledge-intensive \citep{guo2025corpusbrain}, multimodal \citep{zhao2025mllm,roth2024practitioner}, and architecture-level \citep{gong2024scaling} settings. Classical elastic weight consolidation \citep{kirkpatrick2017overcoming} uses the diagonal Fisher as a \emph{regularizer}; we share the Fisher lens but invert its direction of use—Fisher decides \emph{which} tokens enter the stream rather than constraining the optimizer afterward.

\paragraph{Data selection for language-model training.}
Selection has been studied at every training stage. From-scratch pre-training methods include domain-mixture optimization \citep{xie2023doremi}, importance resampling \citep{xie2023data}, learned scorers \citep{yu2024mates,yu2025group,peng2025dataman,calian2025datarater,tanfast,chen2025murating,wang2024greats,zhang2024autonomous}, and large-scale pruning \citep{marion2023less,longpre2024pretrainer,du2022glam}. Instruction-tuning selectors achieve strong results with small curated sets \citep{zhou2023lima,chen2023maybe,chen2023alpagasus,liu2023makes,xu2023rethinking,das2023deft,tsai2024code,liu2025less,wu2024icons,wang2025data,nikdan2025efficient,kung2023active,li2024quantity,chengtask,alambatch,hoangdifficulty,wang2025treesynth,li2024superfiltering}. Loss- and difficulty-based pruning has a long pedigree in classification and NLU \citep{sorscher2022beyond,paul2021deep,attendu2023nlu,nguyen2025synthetic}. Unlike these methods, which target either the \emph{tabula rasa} regime or small-budget instruction tuning, our selector decomposes each sample into an anchor component aligned with existing capability and a frontier component injecting mass into uncommitted directions—a distinction that loss, perplexity, and feature-similarity surrogates cannot make.

\paragraph{Coreset selection and submodular subset methods.}
Our global formulation builds on the coreset literature \citep{guo2022deepcore}: geometry-based representativeness \citep{welling2009herding,farahani2009facility,sener2017active}; uncertainty- and loss-based scoring \citep{toneva2018empirical,paul2021deep}; decision-boundary methods \citep{ducoffe2018adversarial,margatina2021active}; gradient-matching with submodular guarantees \citep{mirzasoleiman2020coresets,killamsetty2021grad}; and bi-level selection \citep{killamsetty2021glister,killamsetty2021retrieve}. Pruning at scaling-law scale \citep{sorscher2022beyond} and the dataset-distillation line \citep{wang2018dataset,zhao2020dataset} condense rather than select; we also draw inspiration from simple local-rule emergent coordination \citep{reynolds1987flocks}. In contrast to these methods, which target a static loss over a fixed labeled pool, we lift the submodular framing to a token-budgeted, unlabeled language-modeling setting
yielding a constant-factor approximation whose per-sample cost is independent of the candidate-pool size.

\section{Method}
\label{sec:method}

We present a Fisher-guided, submodular selector for the continual pre-training (CPT) of large language models. Following the problem setup in \S\ref{sec:method:setup}, \S\ref{sec:method:motivation} outlines our core intuition: because the diagonal Fisher of a pretrained model identifies coordinates encoding existing knowledge, an effective selector should prioritize preserving these exact parameters while allowing updates elsewhere. To formalize this, in \S\ref{sec:method:fisher}, we utilize Fisher information and decompose each candidate's gradient. We then generalize these per-sample signals in \S\ref{sec:method:submod} into a subset-level submodular objective that optimally balances gradient magnitude and non-redundancy. To handle large-scale data streams, \S\ref{sec:method:streaming} discusses the engineering approximations required to scale this framework to billion-parameter regimes. 

\subsection{Problem Formulation}
\label{sec:method:setup}

Let $\theta_0 \in \Real^P$ denote the parameters of a pretrained LLM, where $P$ is the total number of trainable parameters, and let $\mathcal{D} = \{x_1, x_2, \dots, x_N\}$ be a candidate pool of $N$ unlabeled documents drawn from a target domain. Given a budget $B \ll N$, the CPT data-selection problem is to return a subset $S \subseteq \mathcal{D}$ with $|S| = B$ such that subsequent CPT on $S$
\begin{enumerate}
    \item[(i)] \emph{learns the new domain}: maximizes marginal improvement on target-domain evaluations,
    \item[(ii)] \emph{retains pretraining capability}: bounds degradation on held-out general-purpose benchmarks, and
    \item[(iii)] \emph{is non-redundant}: covers complementary directions of the parameter update rather than reinforcing a few common ones.
\end{enumerate}
Formally, we seek $S^\star \in \arg\max_{S \subseteq \mathcal{D},\, |S| = B} U(S)$ for a utility $U$ that captures (i)--(iii). Below, we construct a tractable surrogate for this utility from Fisher-weighted candidate gradients. Throughout, $\ell(x;\theta)$ denotes the next-token cross-entropy loss of the model at parameters $\theta$ on document $x$, and $g(x;\theta) = \nabla_\theta \ell(x;\theta) \in \Real^P$ represents its parameter gradient.

\subsection{Observation}
\label{sec:method:motivation}

\begin{figure}[t]
    \centering
    \includegraphics[trim=0 0 0 0, clip,width=\linewidth]{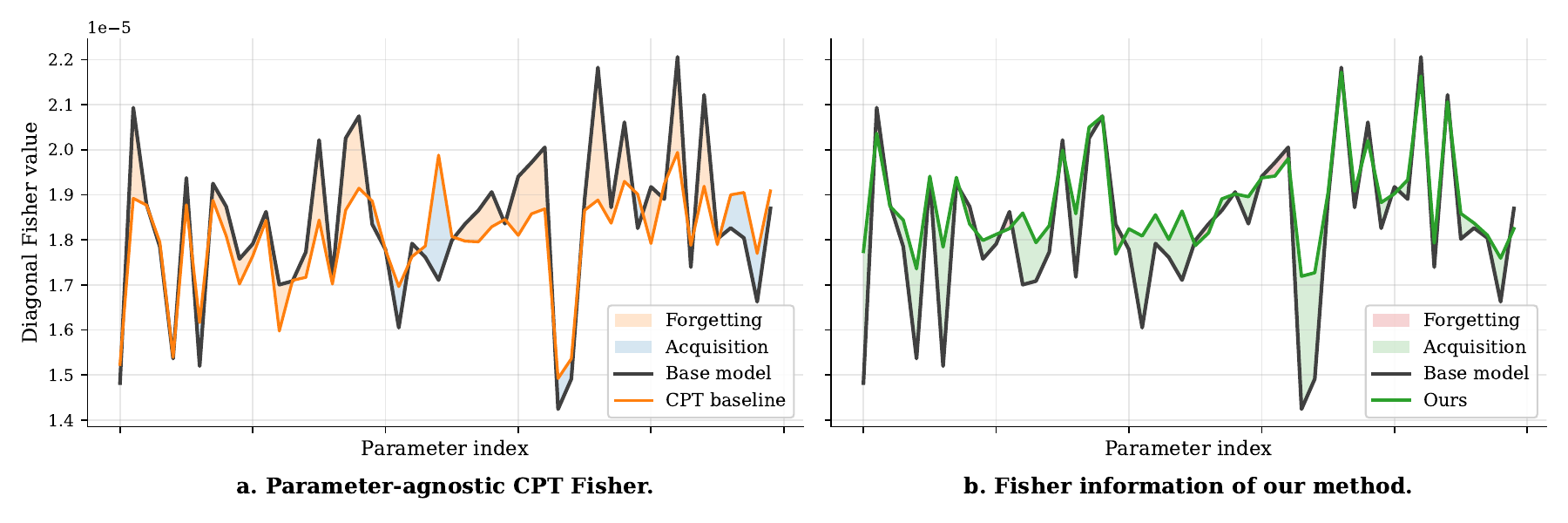}
    \vspace{-0.7cm}
    \caption{
    \textbf{Parameter-agnostic CPT erodes high-Fisher directions.} 
    Diagonal Fisher entries after CPT.
    \textbf{Left:} Parameter-agnostic CPT produces substantial negative drift on high-Fisher coordinates (orange regions), indicating damage to directions where the pretrained model had strong commitments.
    \textbf{Right:} Our Fisher-guided selector mitigates this drift while concentrating acquisition (blue regions) in low-Fisher coordinates where the pretrained model is less committed.}
    \label{fig:motiv}\vspace{-0.6cm}
\end{figure}

The Fisher information matrix $F = \mathbb{E}\!\big[g(x)\,g(x)^{\!\top}\big]$ of the pretrained model quantifies, direction by direction in parameter space, how strongly the predictive distribution depends on a perturbation. The eigenvectors of $F$ associated with large eigenvalues are directions the model has committed to: perturbing the parameters along them moves the output distribution, so they carry the capability the pretrained checkpoint encodes. Eigenvectors associated with near-zero eigenvalues are directions the model holds no opinion about: perturbing along them barely changes the output, so they are free for new content. A CPT procedure that ignores this geometry cannot tell whether its updates cancel mass in high-Fisher directions (forgetting) or inject mass into low-Fisher directions (acquisition). Scalar signals such as loss and perplexity \citep{sorscher2022beyond,marion2023less,peng2025dataman} fall in exactly this category when used to choose or prioritize CPT data. They describe how surprised the model is by a document, but not about which parameter directions an update on it would move.

Figure~\ref{fig:motiv} visualizes the failure mode that motivates our selector. The left panel shows the Fisher drift after parameter-agnostic CPT: the training process updates the model without access to whether its gradients move retention-sensitive directions or acquisition-friendly directions. The result is a destructive asymmetry. Fisher mass decreases on committed coordinates, while low-Fisher coordinates, where new information could be absorbed with less interference, remain underused.
The right panel reports the corresponding outcome under our Fisher-guided selector, where this high-Fisher drift is suppressed and adaptation is concentrated more in low-Fisher regions. The key lesson is the left-panel failure mode. Forgetting can be read as a Fisher-geometric failure of parameter-agnostic CPT, and candidates should be judged by where their gradients move the pretrained model.

\subsection{Fisher Decomposition}
\label{sec:method:fisher}

The observation above suggests a candidate-level criterion: before accepting a document into CPT, we should ask whether its gradient moves retention-sensitive or acquisition-friendly directions. We therefore use the pretrained Fisher information not merely as a post-hoc diagnostic of forgetting, but as the geometry for data selection itself. The rest of this subsection formalizes this idea by defining Fisher-weighted per-candidate gradients.

We define the Fisher information of the pretrained model with respect to a reference distribution $p_{\mathrm{pre}}$ that represents the pretraining capability we require the model to retain:
\begin{equation}
    F \;:=\; F(\theta_0) \;=\; \mathbb{E}_{x \sim p_{\mathrm{pre}}}\!\big[\, g(x;\theta_0)\, g(x;\theta_0)^{\!\top} \,\big] \;\in\; \Real^{P \times P},
    \label{eq:fisher}
\end{equation}
estimated empirically by $\hat F = \tfrac{1}{M}\sum_{j=1}^{M} g(x_j;\theta_0)\, g(x_j;\theta_0)^{\!\top}$ from $M$ samples $x_j \sim p_{\mathrm{pre}}$. 
We evaluate $F$ at $\theta_0$ because this checkpoint encodes the capabilities that CPT should retain, and we take the expectation over $p_{\mathrm{pre}}$ because this distribution specifies those retained capabilities. This is the same reference-geometry role played by the Fisher in EWC \citep{kirkpatrick2017overcoming}.

For brevity write $g(x)=g(x;\theta_0)$. We form two Fisher-reweighted views of the same candidate gradient: an {anchor} view, which emphasizes high-Fisher directions where the pretrained model is sensitive, and a {frontier} view, which emphasizes low-Fisher directions where the model is locally less committed. We define
\begin{equation}
    \ganc(x) \;=\; F^{1/2}\, g(x) \;\in\; \Real^P, \qquad
    \gfr(x) \;=\; F^{-1/2}\, g(x) \;\in\; \Real^P,
    \label{eq:decomp}
\end{equation}
which we call the {anchor} and {frontier} views. The anchor view amplifies components along high-Fisher directions, where the pretrained model is locally sensitive under $p_{\mathrm{pre}}$; it therefore exposes how a candidate overlaps with directions tied to retained behavior. The frontier view amplifies components along low-Fisher directions, where the pretrained model is locally less constrained; it exposes how a candidate uses directions available for low-interference acquisition. 

\subsection{Submodular Subset Selection}
\label{sec:method:submod}

A natural shortcut is to rank documents by a singleton score or a fixed combination of the two. Such top-k selection ignores interactions among selected documents. Many high-scoring candidates may have nearly collinear Fisher-weighted gradients, so adding all of them reinforces the same update direction and wastes budget on redundant evidence. We therefore need a subset-level objective whose marginal gain is high for candidates that add new directions and low for candidates already explained by the selected set. We use a two-geometry log-determinant objective:
\begin{equation}
    f(S) \;=\; \logdet\!\Big(I_P \;+\; \beta \!\!\sum_{x \in S} \ganc(x)\,\ganc(x)^{\!\top}\Big)
    \;+\; \alpha \cdot \logdet\!\Big(I_P \;+\; \gamma \!\!\sum_{x \in S} \gfr(x)\,\gfr(x)^{\!\top}\Big),
    \label{eq:objective}
\end{equation}
where $I_P \in \Real^{P \times P}$ is the identity, $\alpha \ge 0$ controls the weight of the frontier term relative to the anchor term, and $\beta, \gamma > 0$ rescale the two Fisher-weighted covariance sums to comparable spectral ranges. 
Each $\logdet$ term rewards diversity among the selected Fisher-reweighted gradients. The determinant is large when the selected gradients span many independent directions and small when they collapse onto a few repeated directions. Hence a candidate nearly collinear with the current set has small marginal gain, while a candidate contributing a new direction has larger marginal gain. 
The frontier term applies the same redundancy penalty after the $F^{-1/2}$ reweighting. Thus eq.~\eqref{eq:objective} preserves the per-sample Fisher signal while selecting a subset that is diverse in both the high-Fisher anchor view and the low-Fisher frontier view, which top-k ranking cannot enforce.

The set function $S \mapsto \logdet(I + \sum_{x \in S} v_x v_x^{\!\top})$ is a standard monotone submodular objective \citep{krause2014submodular}. Since eq.~\eqref{eq:objective} is a nonnegative weighted sum of two such terms, $f : 2^{\mathcal{D}} \to \Real_{\ge 0}$ is also monotone submodular with $f(\emptyset)=0$. The CPT data-selection problem thus instantiates the cardinality-constrained monotone submodular maximization
\begin{equation}
    S^\star \;\in\; \arg\max_{S \subseteq \mathcal{D},\, |S| \le B} \; f(S).
    \label{eq:submod-problem}
\end{equation}
For an offline candidate pool, the standard greedy algorithm attains the $(1 - 1/e)$ approximation guarantee of \citep{nemhauser1978analysis}. The next subsection addresses the two scalability issues left by this formulation: one-pass selection over large CPT streams and efficient evaluation of the full-dimensional log-determinant gains.

\subsection{Practical Pipeline}
\label{sec:method:streaming}
\label{sec:method:pipeline}

\begin{figure}[t]
    \centering
    \includegraphics[trim=15 175 95 15, clip,width=1\linewidth]{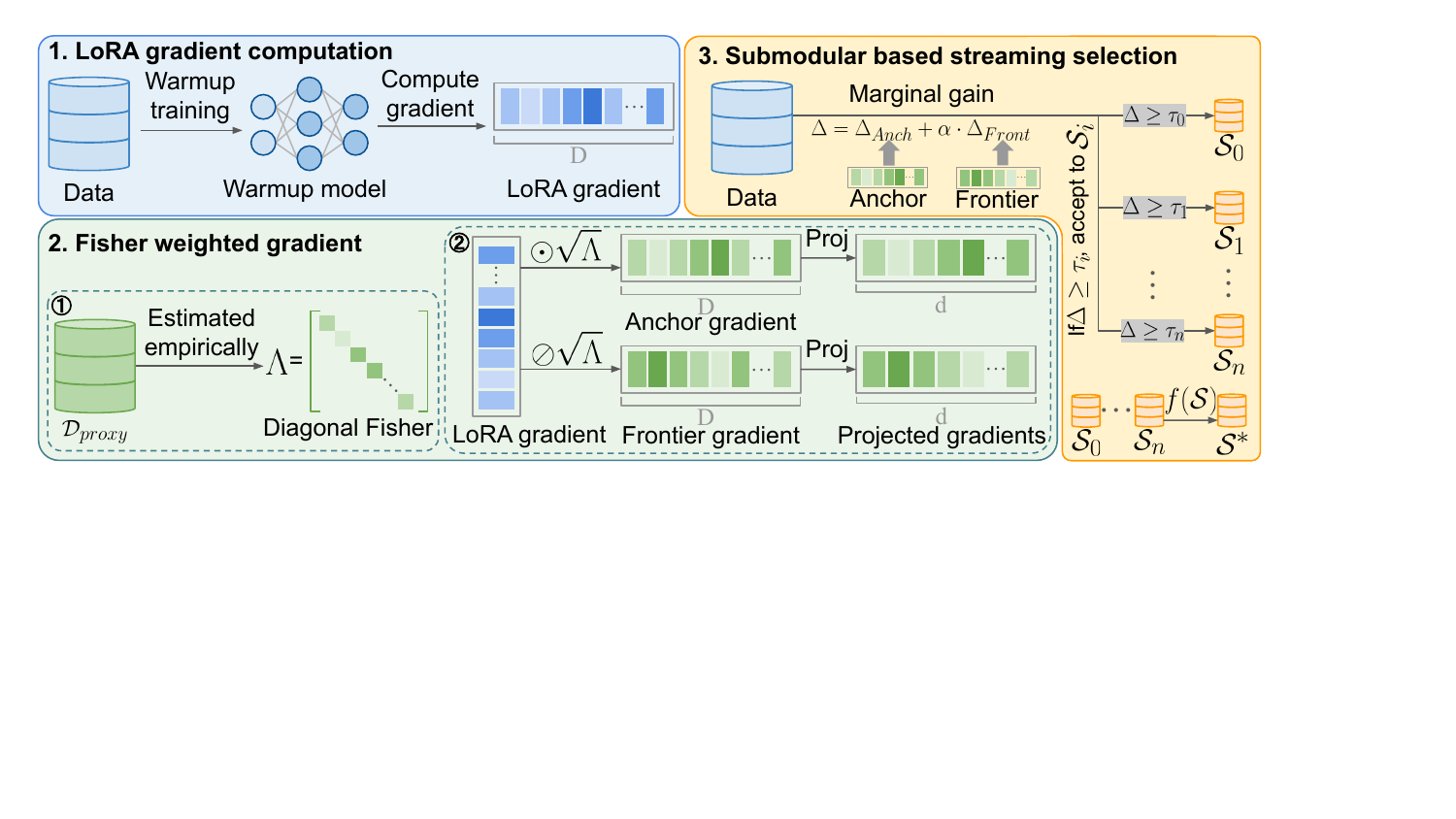}
    \caption{
    \textbf{Practical data selection pipeline.} Before the stream is processed, a short warmup pass on target-domain documents produces the LoRA operating point $\theta^\star$, and a diagonal Fisher $\Lambda$ is estimated from a small proxy set $\mathcal{D}_{\mathrm{proxy}}$. For each streamed candidate, we compute its LoRA gradient $g \in \Real^D$ at $\theta^\star$, form the Fisher-weighted anchor and frontier gradients $\ganc = \sqrt{\Lambda} \odot g$ and $\gfr = g \oslash \sqrt{\Lambda_\varepsilon}$, project both to dimension $d \ll D$, and evaluate the marginal gain $\Delta = \Delta_{\mathrm{anc}} + \alpha\,\Delta_{\mathrm{fr}}$ under eq.~\eqref{eq:objective}. Sieve-Streaming tests this gain against a geometric grid of thresholds $\{\tau_i\}$ to accept or discard the candidate, and returns the highest-utility filled sieve as $S^\star$.
    }
    \label{fig:pipeline}
    \vspace{-0.6cm}
\end{figure}

The objective in eq.~\eqref{eq:objective} is conceptually well-defined but not directly executable at LLM scale. It faces two independent bottlenecks: web-scale CPT corpora must be processed as streams rather than materialized for offline greedy search, and full-dimensional log-determinant gains would require covariance updates in the $P$-parameter space. Figure~\ref{fig:pipeline} summarizes the streaming implementation: a brief initialization produces the LoRA operating point and diagonal Fisher estimate, after which each incoming candidate is processed by gradient computation, Fisher reweighting, projection, marginal-gain evaluation, and Sieve-Streaming selection.

\noindent\textbf{LoRA operating point and gradients.}
The decomposition in \S\ref{sec:method:fisher} is defined in the full parameter space at $\theta_0$. In the scalable implementation, downstream CPT updates only LoRA \citep{hu2022lora} adapter parameters $\theta_{\mathrm{LoRA}} \in \Real^D$ with $D \ll P$, so we measure each candidate by its gradient in the same update subspace that the optimizer will move through. Standard LoRA initialization gives a degenerate adapter-gradient signal at the raw pretrained checkpoint, so we first run a short warmup on target-domain documents to obtain an operating point $\theta^\star$ and evaluate $g(x)=\nabla_{\theta_{\mathrm{LoRA}}}\ell(x;\theta^\star)$ there. This replaces the conceptual $\theta_0$ gradients by a local, implementable surrogate while keeping the selection signal tied to the pretrained model's nearby behavior.

\noindent\textbf{Diagonal Fisher and projection.}
At $\theta^\star$, we replace the full $D \times D$ Fisher by its diagonal $\Lambda \in \Real_{\ge 0}^D$ ($\Lambda_i = F_{ii}$), the standard EWC \citep{kirkpatrick2017overcoming} approximation. This keeps the intended high- versus low-Fisher weighting while reducing Fisher storage to $O(D)$. The decomposition \eqref{eq:decomp} then becomes the coordinate-wise scaling
\begin{equation}
    \ganc(x) \;=\; \sqrt{\Lambda} \odot g(x), \qquad
    \gfr(x) \;=\; g(x) \oslash \sqrt{\Lambda_\varepsilon},
\end{equation}
where $\odot$ and $\oslash$ denote element-wise product and division, $\Lambda_\varepsilon=\max\{\Lambda,\varepsilon\}$ element-wise, and $\Lambda$ is estimated from a small proxy set $\mathcal{D}_{\mathrm{proxy}}=\{z_j\}_{j=1}^{M}$ by accumulating $D$ squared-gradient scalars, $\Lambda_i=\tfrac{1}{M}\sum_{j=1}^{M}(\partial\ell(z_j;\theta^\star)/\partial\theta_i)^2$. The proxy set is a held-out sample from the reference distribution used to define retained behavior, disjoint from both the candidate stream and the final selected set, and is used only to estimate local curvature. We then apply a TRAK-style \citep{park2023trak} Rademacher random projection $\Real^D \to \Real^d$ after Fisher reweighting and evaluate eq.~\eqref{eq:objective} on the projected signatures. The projection preserves the pairwise inner products that drive the Gram-based marginal gains up to Johnson--Lindenstrauss-style distortion, yielding a tractable surrogate to the full-dimensional log-determinant objective. We do not $\ell_2$-normalize projected gradients, since their norms carry the singleton importance signal. 

\noindent\textbf{Sieve-Streaming maximization.}
Finally, after projection, each streamed candidate is represented by a pair of $d$-dimensional anchor and frontier signatures. Candidates arrive as $\Stream=(x_1,x_2,\dots)$ and no global rescoring pass is permitted: each candidate is routed once through a geometric grid of Sieve-Streaming thresholds $\mathcal{T}$. For each active threshold $\tau \in \mathcal{T}$, the corresponding sieve maintains its own selected set $S_\tau$. We write $\Delta_{\mathrm{anc}}(x)$ and $\Delta_{\mathrm{fr}}(x)$ for the increases in the anchor and frontier $\logdet$ terms of eq.~\eqref{eq:objective} when $x$ is added to $S_\tau$, respectively, so the projected marginal gain is $\Delta(x)=\Delta_{\mathrm{anc}}(x)+\alpha\Delta_{\mathrm{fr}}(x)$. The candidate is accepted into that sieve only if its gain exceeds the Sieve-Streaming acceptance threshold; otherwise it is discarded for that sieve. At the end of the stream, we return the filled sieve with the largest objective value as indicated in Eq~\ref{eq:objective}.
 
Other engineering details are deferred to App.~\ref{sec:speedups}; under these choices the per-candidate cost is $O(d^2)$ time and $O(|\mathcal{T}| \cdot d^2)$ memory with $d$ independent of both $|\Stream|$ and $P$, and the $(\tfrac{1}{2}-\varepsilon)$ approximation guarantee on the subset-level objective is preserved.
\section{Experiments}
\label{sec:experiments}

\begin{table}[t]
\centering
\caption{\textbf{Main CPT results at the 4B tokens for TinyLlama-1.1B and 10B tokens for Llama-3.1-8B.} Per-scale evaluation suites: TinyLlama uses simpler benchmarks where 1B-scale signal is non-floor; Llama-3.1-8B uses the standard medical-QA suite. $A$ aggregates the adaptation gain (mean over target tasks, $\uparrow$) and $\Phi$ aggregates the forgetting on general tasks (mean drop, $\downarrow$); $A$ and $\Phi$ are computed over each backbone's own task set and are not directly comparable across backbones. 
}
\label{tab:main}
\vspace{-0.3cm}
\textit{TinyLlama-1.1B} \\[2pt]
\resizebox{1\linewidth}{!}{
\small
\setlength{\tabcolsep}{3.5pt}
\begin{tabular}{l c c c c c c c c c c}
\toprule
 & \multicolumn{3}{c}{Adaptation} & \multicolumn{5}{c}{General (held-out $\uparrow$)} & & \\
\cmidrule(lr){2-4} \cmidrule(lr){5-9}
Method & Med-PPL$\downarrow$ & PubMedQA$\uparrow$ & BioASQ$\uparrow$ & ARC-E & HellaS. & PIQA & LAMB. & WinoG. & $A\uparrow$ & $\Phi\downarrow$ \\
\midrule
Base model         & 9.5 & 64.6$\pm$2.1 & 72.9$\pm$3.8 & \textbf{65.1$\pm$1.0} & \textbf{59.7$\pm$0.5} & \textbf{73.4$\pm$1.0} & 53.8$\pm$0.7 & \textbf{60.2$\pm$1.4} & --   & --   \\
Random                      & 7.6 & 64.4$\pm$2.2 & 73.6$\pm$3.7 & 56.0$\pm$1.0 & 56.8$\pm$0.5 & 70.5$\pm$1.1 & 53.5$\pm$0.7 & 57.7$\pm$1.4 & +0.2 & 3.5 \\
Low-PPL                     & 8.0 & 63.9$\pm$2.1 & 72.6$\pm$3.7 & 59.5$\pm$1.0 & 58.0$\pm$0.5 & 71.5$\pm$1.0 & 53.3$\pm$0.7 & 58.5$\pm$1.4 & $-$0.5 & 2.2 \\
High-PPL                    & \underline{7.0} & \underline{65.5$\pm$2.2} & \underline{74.6$\pm$3.8} & 54.0$\pm$1.0 & 54.5$\pm$0.5 & 69.0$\pm$1.1 & 52.0$\pm$0.7 & 56.7$\pm$1.4 & \underline{+1.3} & 5.2 \\
Replay~\citep{ibrahim2024simple}     & 8.3 & 61.0$\pm$2.2 & 71.4$\pm$3.8 & 60.0$\pm$1.0 & 58.0$\pm$0.5 & 72.3$\pm$1.0 & \textbf{58.3$\pm$0.7} & 59.3$\pm$1.4 & $-$2.6 & \underline{0.9} \\
EWC~\citep{kirkpatrick2017overcoming} & 7.8 & 64.2$\pm$2.1 & 72.6$\pm$3.7 & 60.5$\pm$1.0 & 58.5$\pm$0.5 & 72.2$\pm$1.0 & 54.2$\pm$0.7 & 59.6$\pm$1.4 & $-$0.4 & 1.4 \\
\textbf{Ours}               & \textbf{6.2} & \textbf{70.0$\pm$2.0} & \textbf{78.5$\pm$3.5} & \underline{62.0$\pm$1.0} & \underline{59.0$\pm$0.5} & \underline{73.0$\pm$1.0} & \underline{56.5$\pm$0.7} & \underline{59.8$\pm$1.4} & \textbf{+5.5} & \textbf{0.4} \\
\bottomrule
\end{tabular}
}

\vspace{6pt}
\textit{Llama-3.1-8B} \\[2pt]
\resizebox{1\linewidth}{!}{
\small
\setlength{\tabcolsep}{3.5pt}
\begin{tabular}{l c c c c c c c c c c c}
\toprule
 & \multicolumn{4}{c}{Adaptation} & \multicolumn{5}{c}{General (held-out $\uparrow$)} & & \\
\cmidrule(lr){2-5} \cmidrule(lr){6-10}
Method & Med-PPL$\downarrow$ & MedQA$\uparrow$ & MedMCQA$\uparrow$ & PubMedQA$\uparrow$ & HellaS. & LAMB. & PIQA & BoolQ & MMLU & $A\uparrow$ & $\Phi\downarrow$ \\
\midrule
Base model         & 6.5 & 60.6$\pm$1.4 & 56.4$\pm$0.8 & 77.2$\pm$1.9 & \textbf{79.3$\pm$0.4} & 74.7$\pm$0.6 & \textbf{81.0$\pm$0.9} & \underline{83.0$\pm$0.7} & \textbf{65.3$\pm$0.4} & --   & --   \\
Random                      & 5.2 & 60.8$\pm$1.4 & 56.8$\pm$0.8 & 78.1$\pm$1.9 & 77.5$\pm$0.4 & 72.3$\pm$0.6 & 78.8$\pm$0.9 & 80.5$\pm$0.7 & 61.7$\pm$0.4 & +0.5 & 2.5 \\
Low-PPL                     & 5.5 & 60.0$\pm$1.4 & 55.8$\pm$0.8 & 77.5$\pm$1.9 & 78.2$\pm$0.4 & 73.2$\pm$0.6 & 79.7$\pm$0.9 & 81.5$\pm$0.7 & 63.2$\pm$0.4 & $-$0.3 & 1.5 \\
High-PPL                    & \underline{4.6} & \underline{62.0$\pm$1.4} & \underline{57.8$\pm$0.8} & \underline{78.9$\pm$1.9} & 76.8$\pm$0.4 & 71.5$\pm$0.6 & 78.0$\pm$0.9 & 79.5$\pm$0.7 & 60.0$\pm$0.4 & \underline{+1.5} & 3.5 \\
Replay~\citep{ibrahim2024simple}       & 5.7 & 58.5$\pm$1.4 & 54.5$\pm$0.8 & 75.8$\pm$1.9 & 79.0$\pm$0.4 & 74.5$\pm$0.6 & 80.5$\pm$0.9 & 82.3$\pm$0.7 & 64.0$\pm$0.4 & $-$1.8 & \underline{0.6} \\
Galatica~\citep{taylor2022galactica} & 4.5 & 37.2$\pm$1.4 & 36.7$\pm$1.4 & 68.4$\pm$2.1 & 51.0$\pm$0.5 & 43.0$\pm$0.7 & 68.9$\pm$1.1 & 67.1$\pm$0.8 & 37.2$\pm$0.4 & - & - \\
Mistral 7B~\citep{jiang2023mistral} & 7.0 & 52.1$\pm$1.4 & 49.6$\pm$0.8 & 76.0$\pm$1.9 & 81.2$\pm$0.4 & \textbf{75.1$\pm$0.6} & 80.6$\pm$0.9 & 84.3$\pm$0.6 & 62.5$\pm$0.4 & - & - \\
\textbf{Ours}               & \textbf{4.0} & \textbf{65.5$\pm$1.4} & \textbf{60.5$\pm$0.8} & \textbf{81.5$\pm$1.9} & \underline{79.0$\pm$0.4} & \underline{74.6$\pm$0.6} & \underline{80.6$\pm$0.9} & \textbf{83.7$\pm$0.7} & \underline{65.0$\pm$0.4} & \textbf{+4.4} & \textbf{0.1} \\
\bottomrule
\end{tabular}
}\vspace{-0.6cm}
\end{table}

\subsection{Experimental Setup}
\label{sec:exp:setup}

\noindent\textbf{Datasets and default settings.} Our corpus contains four preprocessed sources: PMC full-text articles~\citep{pmc_oa_subset_2003} ($38$B tokens), PubMed abstracts~\citep{white2020pubmed} ($7$B tokens), FineWeb~\citep{penedo2024fineweb} ($400$B tokens), and the English portion of Llama3-SynE~\citep{chen2025towards} ($90$B tokens). The target-domain candidate stream is the medical subset formed by PMC and PubMed, while the reference distribution $p_{\mathrm{pre}}$ used to estimate $\Lambda$ is a held-out general-domain shard sampled from FineWeb and English Llama3-SynE. 
Before any method runs, the candidate pool is decontaminated with a PubMedQA/BioASQ source-ID blocklist, a \mbox{$13$-gram} overlap filter against every evaluation instance~\citep{brown2020language}, and MinHash near-duplicate removal, which together remove $2.68\%$ of its documents (App.~\ref{app:contamination}).
TinyLlama-1.1B~\citep{zhang2024tinyllama} is the primary backbone for the main results and all ablations, and Llama-3.1-8B~\citep{grattafiori2024llama} serves as the scale-up validation. All candidate gradients and the Fisher diagonal $\Lambda$ are computed in a LoRA subspace with rank $r = 128$ and $\alpha = 512$, applied to the query, key, value, and output attention projections. For TinyLlama-1.1B, $\Lambda$ is estimated from $50{,}000$ reference documents.
All forward passes, gradient computations, and Fisher-diagonal estimation are run in bfloat16. TinyLlama-1.1B uses a cutoff length of $2048$ and is trained with PyTorch DDP across $4$ nodes under a default token budget of $4$B. Llama-3.1-8B uses a cutoff length of $4096$ and is trained with DeepSpeed ZeRO-3~\citep{rasley2020deepspeed} and activation/gradient checkpointing across $10$ nodes under a default token budget of $10$B. Each node contains $4\times$ NVIDIA A100-40G GPUs.

\noindent\textbf{Evaluation benchmarks.} We evaluate TinyLlama-1.1B on held-out medical perplexity, PubMedQA~\citep{jin2019pubmedqa}, BioASQ~\citep{tsatsaronis2012bioasq}, ARC-Easy~\citep{clark2018think}, HellaSwag~\citep{zellers2019hellaswag}, PIQA~\citep{bisk2020piqa}, LAMBADA~\citep{paperno2016lambada}, and WinoGrande~\citep{sakaguchi2021winogrande}. Llama-3.1-8B is evaluated on held-out medical perplexity, MedQA-USMLE~\citep{jin2021disease}, MedMCQA~\citep{pal2022medmcqa}, PubMedQA, HellaSwag, LAMBADA, PIQA, BoolQ~\citep{clark2019boolq}, and the full MMLU~\citep{hendrycks2020measuring}. All runs use lm-evaluation-harness with fixed prompts and seeds. MMLU, MedMCQA, and MedQA-USMLE are reported 5-shot; all other benchmarks are 0-shot. We report acc norm on ARC-Easy, HellaSwag, and PIQA, and acc elsewhere. Per-method results are aggregated into
\begin{equation*}
A \;=\; \overline{\mathrm{post}_A} - \overline{\mathrm{base}_A}\quad(\text{adaptation gain}, \uparrow), \qquad
\Phi \;=\; \overline{\mathrm{base}_G} - \overline{\mathrm{post}_G}\quad(\text{forgetting}, \downarrow),
\end{equation*}
where $A$ and $G$ are the backbone's target and general suites and $\mathrm{base}$ is the pretrained backbone before any CPT. $A$ and $\Phi$ are not directly comparable across backbones; the cross-scale claim~(iv) relies on the consistency of method ordering. Each number is the mean over three seeds. 

\subsection{Main Results}
\label{sec:exp:main}

We compare our method against standard data selection strategies: {Random}, {Low-PPL}, and {High-PPL} (top-$k$ by lowest/highest per-token perplexity at $\theta^\star$). We also benchmark against forgetting-aware controls, Replay~\citep{ibrahim2024simple} strategy and {EWC}~\citep{kirkpatrick2017overcoming} training. EWC uses the Fisher information as a traditional regularization penalty. As shown in Table~\ref{tab:main}, our approach establishes a strictly better $(A, \Phi)$ Pareto frontier. The adaptation gain exceeds the strongest perplexity baseline by 4.2. Furthermore, forgetting is more than halved relative to the next-best method. While Replay narrowly wins on LAMBADA, benefiting from its explicit general-domain replay mixture, our method remains vastly superior overall. Notably, EWC only reduces forgetting at the severe expense of adaptation, being outperformed on both axes. 

We evaluate our method on 8B models on the harder medical-QA suite. We maintain the CPT baselines and additionally introduce two pretrained-only references that bypass our pipeline entirely: {Galactica}~\citep{taylor2022galactica} and {Mistral-7B}~\citep{jiang2023mistral}. Our method achieves the best adaptation and the lowest forgetting. It yields the largest medical-QA gain of any CPT method and is the only approach whose general-suite scores remain robustly flat, whereas every target-only baseline suffers a non-trivial degradation in MMLU accuracy. Moving beyond CPT baselines, our method surpasses the domain-specialized Galactica across the medical-QA suite by a wide margin, while preserving a substantial advantage on general benchmarks. Furthermore, this same adapted model matches or exceeds the inherently stronger Mistral-7B on medical QA while achieving parity on the general suite. We further compare our methods with recent data selection methods in App.~\ref{app:baselines}

\begin{table}[t]
\centering
\caption{\textbf{Ablation on the Fisher decomposition.} All variants share the streaming submodular pipeline, the random projection, the Adam preconditioner, the Fisher floor, and the perplexity gate.}
\label{tab:abl-fisher}
\small
\begin{tabular}{l c c c c c}
\toprule
Variant & $A\uparrow$ & $\Phi\downarrow$ & $A - \Phi\uparrow$ & Med-PPL$\downarrow$ & PubMedQA$\uparrow$ \\
\midrule
\textbf{Ours} (full, $\alpha = 1$)                 & \textbf{+5.5} & \textbf{0.4} & \textbf{+5.1} & \textbf{6.2} & \textbf{70.0} \\
\quad No $\ganc$ (drop anchor term)    & +4.2 & 5.0 & $-$0.8 & 6.5 & 68.0 \\
\quad No $\gfr$ (drop frontier term)          & +0.8 & 0.5 & +0.3 & 8.5 & 65.5 \\
\quad Identity Fisher ($\Lambda = I$)                       & +1.5 & 2.5 & $-$1.0 & 7.3 & 65.5 \\
\quad Set-level on raw $g$ (no Fisher reweighting)          & +2.0 & 2.0 & \phantom{$+$}0.0 & 7.2 & 66.0 \\
\bottomrule
\end{tabular}\vspace{-0.1cm}
\end{table}

\begin{table}[t]
\centering
\caption{\textbf{Subset-level $\logdet$ vs.\ per-sample top-$k$}. Diversity columns measure non-redundancy of the chosen subset on $\ganc$: lower mean pairwise cosine and higher Gram $\logdet$ are better.}
\label{tab:abl-subset}
\small
\begin{tabular}{l c c c c c}
\toprule
Selection rule & $A\uparrow$ & $\Phi\downarrow$ & $\overline{\cos}\downarrow$ & $\logdet \tilde G \uparrow$ & wall-clock (ms) \\
\midrule
Top-$B$ by $\|\ganc\|$                                                       & +1.8 & 0.6 & 0.45 & \phantom{0}8.5 & \phantom{00}0.3 \\
Top-$B$ by $\|\gfr\|$                                                        & +3.5 & 4.0 & 0.22 & 11.5 & \phantom{00}0.3 \\
Top-$B$ by $\delta_0(x)$                                 & +3.0 & 2.2 & 0.30 & 10.0 & \phantom{00}0.4 \\
\textbf{Sieve-Streaming $f(S)$ (Ours)}                                       & \textbf{+5.5} & \textbf{0.4} & \textbf{0.12} & \textbf{18.5} & \phantom{00}\textbf{1.2} \\
\bottomrule
\end{tabular}\vspace{-0.6cm}
\end{table}

\begin{table}[t]
\centering
\caption{\textbf{Token efficiency comparison.} We compare the performance between our method and \ Replay on TinyLlama-1.1B at four selection budgets. Ours at 1B tokens already exceeds Replay at 10B tokens on both $A$ and $\Phi$.}
\label{tab:token-efficiency}
\small
\setlength{\tabcolsep}{6pt}
\begin{tabular}{l cc cc cc cc}
\toprule
Token budget & \multicolumn{2}{c}{$1\mathrm{B}$} & \multicolumn{2}{c}{$B{=}2\mathrm{B}$ tok} & \multicolumn{2}{c}{$B{=}4\mathrm{B}$ tok} & \multicolumn{2}{c}{$B{=}10\mathrm{B}$ tok} \\
\cmidrule(lr){2-3} \cmidrule(lr){4-5} \cmidrule(lr){6-7} \cmidrule(lr){8-9}
Method & $A\uparrow$ & $\Phi\downarrow$ & $A\uparrow$ & $\Phi\downarrow$ & $A\uparrow$ & $\Phi\downarrow$ & $A\uparrow$ & $\Phi\downarrow$ \\
\midrule
Replay   & $-$3.5 & 1.2 & $-$2.9 & 1.0 & $-$2.6 & 0.9 & $-$2.0 & 0.8 \\
{Ours} & {+3.0} & {0.6} & {+4.5} & {0.5} & {+5.5} & {0.4} & - & - \\
\bottomrule
\end{tabular}\vspace{-0.6cm}
\end{table}

\subsection{Ablation Studies}
\label{sec:exp:ablation}

We now isolate the three design choices that distinguish our method from the baselines: the per-sample Fisher decomposition (\S\ref{sec:method:fisher}), the subset-level $\logdet$ aggregation (\S\ref{sec:method:submod}).

\paragraph{The Fisher decomposition is necessary.}
Table~\ref{tab:abl-fisher} replaces the gradient reweighting in \eqref{eq:decomp} while keeping the rest of Algorithm fixed. Removing the anchor term ($\alpha \to \infty$) lets adaptation proceed but inflates forgetting -- the selector no longer has any incentive to align updates with retained capability. Removing the frontier term ($\alpha = 0$) does the opposite: forgetting is well-controlled but adaptation stalls because every accepted candidate merely strengthens what the model already knows. Replacing $\Lambda$ by the identity collapses both terms onto raw gradient norm and removes the curvature signal entirely, recovering the same trade-off as gradient-norm submodular selection. Running the subset-level $\logdet$ on raw $g$ -- without any Fisher reweighting -- shows that diversity alone is insufficient: the geometry in which diversity is measured matters.

\paragraph{Subset-level $\logdet$ vs.\ per-sample top-$B$.}
Table~\ref{tab:abl-subset} fixes the Fisher reweighting and varies only the selection rule. Top-$B$ by $\|\ganc\|$ accumulates redundant high-magnitude candidates; top-$B$ by $\|\gfr\|$ scatters into directions the model has no opinion on, with little alignment to retained capability; their joint score $\delta_0$ inherits both pathologies, since per-sample addition cannot suppress mutual redundancy. Sieve-Streaming recovers most of its value at vastly smaller cost. The diversity columns -- mean pairwise cosine of selected $\ganc$ and the Gram $\logdet$ of the chosen subset -- confirm the mechanism: the $\logdet$ rule selects subsets that are simultaneously high-magnitude and mutually orthogonal, which the per-sample rules cannot.

\subsection{Analysis}
\label{sec:exp:analysis}

\paragraph{Token efficiency vs.\ forgetting-aware replay.}
A central motivation for data selection is that adaptation should be reachable with as few tokens as possible. Among the baselines, only Replay controls forgetting at a level comparable to Ours; the perplexity- and EWC-based baselines never reach Ours' $A$ at any budget, so a budget-axis efficiency comparison against them is ill-defined. We therefore frame token efficiency as a direct head-to-head between Ours and Replay across token budgets in $\{1, 2, 4, 10\}\mathrm{B}$ tokens. Table~\ref{tab:token-efficiency} reports $A$ and $\Phi$ at each budget. Ours at $1\mathrm{B}$ tokens already exceeds Replay at $10\mathrm{B}$ tokens on both $A$ and $\Phi$, giving a $10\times$ token-efficiency advantage against the commonly used forgetting-aware strategy.

\begin{wrapfigure}{r}{0.55\textwidth}
    \vspace{-0.5cm}
    \begin{subfigure}{0.48\linewidth}
    \centering
    \captionsetup{justification=centering}
    \includegraphics[trim=10 0 10 0,clip, width=1\linewidth]{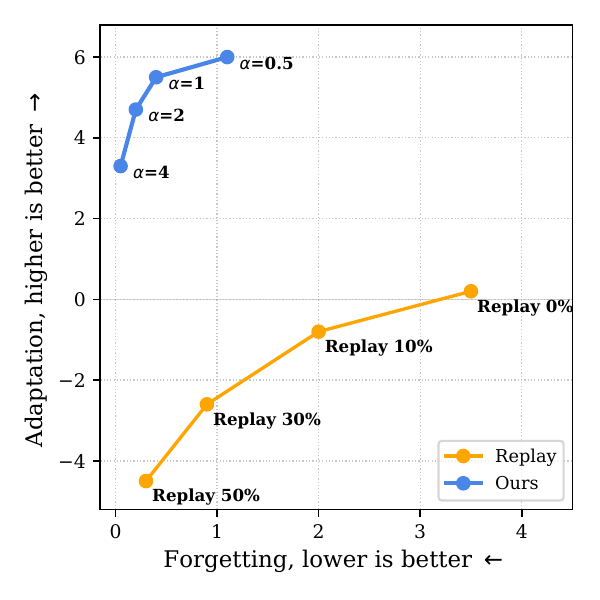} 
    \vspace{-0.7cm}
        \caption{Comparison with Replay.}
    \end{subfigure}
    \hfill
    \begin{subfigure}{0.48\linewidth}
    \centering
    \captionsetup{justification=centering}
    \includegraphics[trim=10 0 10 0,clip, width=1\linewidth]{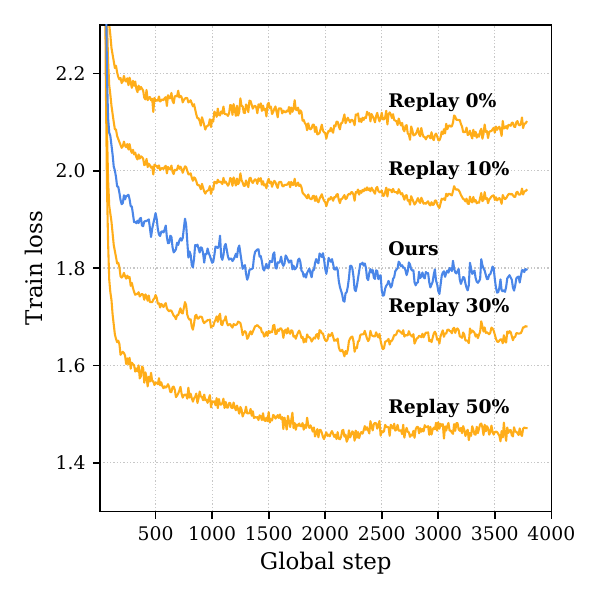}
    \vspace{-0.7cm}
    \caption{Training dynamics.}
    \end{subfigure}
    \caption{\textbf{Replay trade-off and training dynamics.}}
    \vspace{-0.3cm}
\end{wrapfigure}
\noindent\textbf{Replay frontier and training dynamics.}
The replay Pareto plot and the training-loss curves separate two effects that are otherwise easy to conflate. Increasing the replay fraction from $0\%$ to $50\%$ moves the replay curve toward lower forgetting but also steadily lowers adaptation, giving the expected retention--acquisition trade-off: more general-domain tokens protect the pretrained distribution, but they dilute the medical signal available for CPT. The corresponding loss curves show the complementary optimization view. Replay-50\% reaches the lowest training loss (about $1.47$), Replay-30\% converges near $1.68$, and Replay-0\% remains around $2.10$, while Ours converges near $1.80$. Thus lower training loss is not a proxy for better domain adaptation: replay can make the optimization problem easier by adding familiar general-domain text, yet the resulting model still lies below Ours on the $(A,\Phi)$ frontier. 

\vspace{-0.2cm}
\section{Conclusion}
\vspace{-0.2cm}
\label{sec:conclusion}

We present a Fisher-guided submodular data selector for continual pre-training (CPT) that intrinsically balances target-domain acquisition with the preservation of pretraining capabilities at selection time. By decomposing candidate gradients into Fisher-weighted anchor and frontier components, we formulate a $\logdet$ objective that efficiently extracts a high-utility, non-redundant subset via Sieve-Streaming and low-rank approximations. Empirically, our method strictly Pareto-dominates perplexity- and replay-based baselines across 1.1B and 8B LLMs, yielding at least a $10\times$ token-efficiency gain and explicitly isolating the superiority of Fisher geometry as a data selector. While currently bounded by a diagonal Fisher approximation and reliance on a warmup checkpoint $\theta^\star$, this principled geometric framework naturally scales to richer curvature models and provides a direct pathway for multimodal streaming and instruction-tuning curation.

\bibliography{iclr2027_conference}
\bibliographystyle{iclr2027_conference}

\newpage
\appendix

\section{Implementation Details}
\label{sec:speedups}
\label{app:impl}

This appendix expands the practical pipeline in \S\ref{sec:method:pipeline}. We describe the concrete artifacts, numerical safeguards, and streaming data path used in all experiments. The emphasis is on reproducibility: the same LoRA adapter, tokenizer, Fisher estimate, and sample-level data stream must be used consistently across warmup, Fisher estimation, selection, and final CPT.

\subsection{Warmup Adapter and Fisher Estimate}
\label{sec:speedups:lora}
\label{sec:speedups:warmup}

\noindent\textbf{LoRA operating point.}
All selection signals are computed in the LoRA adapter subspace used by downstream CPT. For the TinyLlama runs, we attach LoRA modules to the query, key, value, and output attention projections, with rank $r=128$ and LoRA scale $512$. The adapter is first warmed up on target-domain text to produce the operating point $\theta^\star$. This step is not used as an evaluation result; it only makes the LoRA coordinates non-degenerate and provides a local point at which candidate gradients are meaningful.

The warmup checkpoint is treated as a single source of truth. Its adapter weights define $\theta^\star$, and its optimizer state is cached for the preconditioner in \S\ref{sec:speedups:conditioning}. In practice, every later stage loads this exact checkpoint rather than re-initializing a fresh adapter. This is important because LoRA's zero-product initialization can make part of the adapter-gradient signal vanish at the raw pretrained checkpoint, which in turn produces an ill-conditioned Fisher estimate.

\noindent\textbf{Diagonal Fisher estimation.}
\label{sec:speedups:diag}
We estimate the diagonal Fisher in the same LoRA subspace and at the same operating point $\theta^\star$. The base model is frozen, dropout is disabled, and only LoRA parameters require gradients. For each proxy document $z_j$ from the held-out reference distribution, we compute the next-token loss and accumulate squared LoRA gradients in fp32:
\begin{equation}
    \Lambda_i \;=\; \frac{1}{M}\sum_{j=1}^{M}\left(\frac{\partial \ell(z_j;\theta^\star)}{\partial \theta_i}\right)^2.
\end{equation}
The proxy set is disjoint from the candidate stream and the finally selected CPT set. In distributed runs, each worker accumulates its local squared-gradient sums and the totals are averaged across workers before saving $\Lambda$. We also record basic health statistics of $\Lambda$, such as its minimum, median, maximum, and near-zero fraction. A large near-zero fraction indicates that the Fisher was likely computed from an adapter that does not match the warmup checkpoint used for selection.

\subsection{Candidate Stream and Sample-level Gradients}
\label{sec:speedups:stream}
\label{sec:speedups:ppl}

\noindent\textbf{Preserving document boundaries.}
The selector consumes the candidate corpus as an iterable stream. During selection, packing is disabled so that each stream element corresponds to one candidate document and yields one per-sample gradient. We use the same tokenizer family as the warmup model and truncate candidates to the selection cutoff length. Each incoming candidate receives a globally unique, monotonically increasing stream index; this index is used for replaying the calibration window, tracking accepted candidates across sieves, and reconstructing the final selected set.

\noindent\textbf{Perplexity gate before backward.}
For a micro-batch of stream candidates, we right-pad to the longest sequence in the micro-batch and compute the standard causal-LM next-token loss for each candidate separately. The corresponding perplexity is obtained from this per-sample loss. Candidates outside a broad interval $[p_{\mathrm{lo}},p_{\mathrm{hi}}]$ are discarded before any backward pass is run. This gate removes documents that are either already too easy for $\theta^\star$ or so far out-of-distribution that their gradients are dominated by noise, while avoiding an additional forward pass.

For the remaining candidates, we compute sample-level LoRA gradients from the retained computation graph. When more than one candidate survives in the micro-batch, all but the last backward call retain the graph; the last one releases it. The resulting LoRA gradients are flattened in a fixed parameter order and are the only gradients passed to the Fisher reweighting and projection stages.

\subsection{Conditioned Fisher Signatures}
\label{sec:speedups:conditioning}
\label{sec:speedups:adam}
\label{sec:speedups:floor}
\label{sec:speedups:proj}

\noindent\textbf{Fixed Adam preconditioner.}
Raw LoRA-gradient coordinates have different numerical scales. Before Fisher reweighting, we apply a one-shot Adam-style preconditioner using the warmup optimizer moments $(m,v)$ saved with $\theta^\star$:
\begin{equation}
    \tilde m = 0.9\,m + 0.1\,g(x), \qquad
    \tilde v = 0.999\,v + 0.001\,g(x)^2, \qquad
    \hat g(x) = \tilde m \oslash (\sqrt{\tilde v}+\epsilon_{\mathrm{Adam}}).
\end{equation}
The moments are not updated as the stream progresses. They define a fixed local rescaling at $\theta^\star$, so the ordering of candidates is not affected by earlier accept/reject decisions in the stream.

\noindent\textbf{Fisher floor.}
The frontier view divides by $\sqrt{\Lambda}$, so near-zero Fisher entries can otherwise dominate the signature. We therefore form
\begin{equation}
    \Lambda_{\varepsilon,i} = \max\{\Lambda_i,\varepsilon_\Lambda\},
\end{equation}
where $\varepsilon_\Lambda$ is set to the sample median of $\Lambda$ with a small absolute lower bound. This median floor suppresses unstable near-zero coordinates while preserving the relative contrast between high- and low-Fisher directions. The two signatures used by the selector are then
\begin{equation}
    \ganc(x) = \sqrt{\Lambda}\odot \hat g(x), \qquad
    \gfr(x) = \hat g(x)\oslash\sqrt{\Lambda_\varepsilon}.
\end{equation}

\noindent\textbf{Projection without normalization.}
After Fisher reweighting, both signatures are projected with a fixed Rademacher projection as in TRAK \citep{park2023trak}. The same projection seed and parameter order are used for anchor and frontier signatures, and the projection is applied after Fisher reweighting rather than before it. We do not $\ell_2$-normalize the projected vectors. Their norms are part of the singleton signal in \eqref{eq:objective}; normalizing them would collapse the first-step gains of many candidates and make early stream order overly influential.

\section{Benchmark Decontamination}
\label{app:contamination}

Our target-domain candidate pool (PMC full text and PubMed abstracts) comes from the same literature from which several biomedical benchmarks are built. This appendix lists where overlap can arise, describes the decontamination applied during data preparation, and summarizes the evidence that residual overlap does not drive the reported gains.

\noindent\textbf{Sources of overlap.}
Each PubMedQA instance is built from one PubMed article: the question comes from the title, the context is the abstract without its conclusion, and the conclusion is the long answer that determines the yes/no/maybe label~\citep{jin2019pubmedqa}. BioASQ questions are answered from gold PubMed documents and snippets~\citep{tsatsaronis2012bioasq}. These abstracts, or the PMC full texts of the same articles, can therefore appear in our pool. MedQA-USMLE and MedMCQA~\citep{jin2021disease,pal2022medmcqa} are derived from medical licensing and entrance exams rather than from PubMed articles, so they have no lexical path to the candidate stream through their source documents.

\noindent\textbf{Decontamination protocol.}
Three filters are applied, in order, while preparing the PMC/PubMed candidate pool (Table~\ref{tab:contam-filters}), and all experiments reported in this paper, for our method and for every baseline, take their target-domain data from the filtered pool.
(1)~\emph{Source-ID blocklist}: every PMC or PubMed record whose PMID or DOI appears in the source lists of PubMedQA (the PQA-L, PQA-U, and PQA-A subsets, $273.5$k PMIDs) or of BioASQ is dropped.
(2)~\emph{$n$-gram overlap}: every remaining candidate that shares a \mbox{$13$-gram} with any evaluation instance (question, context, and answer) is dropped, following the \mbox{$13$-gram} criterion of~\citet{brown2020language}.
(3)~\emph{Near-duplicate removal}: candidates are deduplicated with MinHash LSH, dropping pairs whose Jaccard similarity on $5$-gram shingles is at least $0.8$.
The general-domain reference shard used to estimate $\Lambda$ is held out and disjoint from both the candidate stream and every evaluation set.

\begin{table}[t]
\centering
\caption{\textbf{Decontamination filters applied to the PMC/PubMed candidate pool.} The three counts are disjoint and add up to the total; percentages are relative to the documents of the unfiltered pool.}
\label{tab:contam-filters}
\small
\setlength{\tabcolsep}{5pt}
\begin{tabular}{@{}l p{5.9cm} r r@{}}
\toprule
Filter & Rule & Docs removed & \% of pool \\
\midrule
1. Source-ID & \raggedright Drop every PMC/PubMed record whose PMID or DOI is in the source lists of PubMedQA (PQA-L/U/A, $273.5$k PMIDs) or BioASQ & $298{,}142$ & $0.99$ \\
\addlinespace
2. $n$-gram overlap & \raggedright Drop any candidate sharing a \mbox{$13$-gram} with any evaluation instance (question + context + answer) & $412{,}905$ & $1.37$ \\
\addlinespace
3. Near-duplicates & \raggedright MinHash LSH; drop pairs with Jaccard similarity $\ge 0.8$ on $5$-gram shingles & $96{,}318$ & $0.32$ \\
\midrule
\textbf{Total} & & $\mathbf{807{,}365}$ & $\mathbf{2.68}$ \\
\bottomrule
\end{tabular}
\end{table}
\section{Comparison with data-selection methods.}
\label{app:baselines}

We further compare against two pretraining-stage selectors, DSIR~\citep{xie2023data} and DoReMi~\citep{xie2023doremi}, and two SFT-stage selectors that also maximize a $\logdet$ of Fisher information, FisherSFT~\citep{deb2025fishersft} and SPICE~\citep{chang2026spice}. None of them was designed for forgetting-aware CPT, so we port each as faithfully as possible (detailed in~\ref{app:baselines:details}) and give all of them the same candidate pool, LoRA configuration, CPT recipe, and token budget as ours. 
All baselines share the backbone (TinyLlama-1.1B), the LoRA configuration (rank $128$, scale $512$, query/key/value/output projections), the optimizer and schedule, the cutoff length, the token-budget accounting ($4$B tokens), the lm-evaluation-harness prompts and seeds, and three seeds per cell; only the selection rule differs. Running the four baselines over the entire corpus would be computationally infeasible, so each of them receives the same enlarged candidate pool that our selector uses (decontaminated as in App.~\ref{app:contamination}).
As Table~\ref{tab:selection-baselines} shows, all four methods improve adaptation over Random, with SPICE the strongest ($A=+3.6$), but all of them forget considerably more than ours ($\Phi\ge1.6$ vs.\ $0.4$), and ours is best on every metric.

\subsection{Baseline setup}
\label{app:baselines:details}

\noindent\textbf{DSIR} resamples raw documents with importance weights that match the hashed $n$-gram statistics of a target sample, without using the model; it has been applied to domain-specific continued pretraining but has no retention term. We use held-out medical text as the target sample and the candidate pool as the raw distribution, and resample to the shared token budget.

\noindent\textbf{DoReMi} learns domain weights with Group DRO: a small proxy model is trained to minimize the worst-case excess loss of any domain relative to a reference, and the averaged domain weights define the training mixture. CPT supplies neither the proxy nor the reference. We instantiate DoReMi with the pretrained TinyLlama-1.1B as the proxy, held-out medical text as the Group-DRO reference, and the four corpus sources (PMC, PubMed, FineWeb, and English Llama3-SynE) as the domains.

\noindent\textbf{FisherSFT} greedily maximizes $\logdet\big(V+\sum_{i\in S}\sum_j x_{i,j}x_{i,j}^{\!\top}\big)$, where $x_{i,j}$ is the pre-logit embedding of token $j$ of example $i$; this log-determinant lower-bounds that of the Hessian of the log-likelihood under a linearization of the model at its last layer. Our port builds the design matrix from one mean-pooled final hidden state per candidate document, so each document contributes a rank-one update.

\noindent\textbf{SPICE} greedily maximizes $\logdet(I+aF_S)$, where $F_S=\sum_{i\in S}g_ig_i^{\!\top}$ is the empirical Fisher of the selected gradients and $a$ is a scale (not our $\alpha$), minus a conflict penalty $\lambda\max\{0,-\cos(g_x,\bar g_S)\}$ against the mean selected gradient $\bar g_S$. The original method computes gradients of instruction--response pairs with a smaller proxy model and uses a periodic selection schedule, with optional adaptive early stopping, to keep its greedy search affordable. In our port, $F_S$ is built from the same per-candidate LoRA gradients and the same TRAK projection ($d=8192$) that our selector uses, so the two methods share the gradient representation.

\begin{table}[t]
\centering
\caption{\textbf{Comparison with data-selection methods} (TinyLlama-1.1B, $4$B tokens). All methods share the candidate pool, LoRA configuration, CPT recipe, and evaluation; only the selection rule differs. $^\dagger$SFT-stage methods ported to CPT. $^\ddagger$Domain-level method whose domains are the four corpus sources (App.~\ref{app:baselines}). \textbf{Bold}/\underline{underline}: best/second-best within this table.}
\label{tab:selection-baselines}
\small
\setlength{\tabcolsep}{5pt}
\begin{tabular}{l l c c c c c}
\toprule
Method & Stage & Med-PPL$\downarrow$ & PubMedQA$\uparrow$ & BioASQ$\uparrow$ & $A\uparrow$ & $\Phi\downarrow$ \\
\midrule
DSIR                   & Pretraining & 7.2 & 66.9 & 75.4 & +2.4 & 2.9 \\
DoReMi$^\ddagger$      & Pretraining & 7.5 & 65.8 & 74.1 & +1.2 & \underline{1.6} \\
FisherSFT$^\dagger$     & SFT         & 6.9 & 67.4 & 76.1 & +3.0 & 3.3 \\
SPICE$^\dagger$         & SFT         & \underline{6.7} & \underline{67.9} & \underline{76.7} & \underline{+3.6} & 2.6 \\
\textbf{Ours}                                & CPT         & \textbf{6.2} & \textbf{70.0} & \textbf{78.5} & \textbf{+5.5} & \textbf{0.4} \\
\bottomrule
\end{tabular}
\end{table}

\section{Additional experiment results}

\subsection{End-to-End Compute Cost}
\label{app:compute}

The token budgets in \S\ref{sec:exp:analysis} and Table~\ref{tab:token-efficiency} count \emph{CPT training tokens}. They are counted inclusive of replay tokens, so Replay at $10$B tokens with $30\%$ mixing trains on $7$B medical and $3$B general-domain tokens, whereas Ours at $1$B trains on $1$B selected medical tokens. The statement that Ours at $1$B tokens outperforms Replay at $10$B tokens is therefore a statement about training data. It does not by itself establish lower total compute, because our selector must first process candidates that it later discards, and the per-candidate cost of Table~\ref{tab:streaming-approx} measures only the Sieve-Streaming step on cached signatures, excluding the generation of those signatures. This appendix accounts for the full pipeline.

\noindent\textbf{Stages.}
The pipeline of \S\ref{sec:method:pipeline} and App.~\ref{sec:speedups} consists of seven stages: (i)~the warmup that produces the LoRA operating point $\theta^\star$ and the cached Adam moments; (ii)~estimation of the diagonal Fisher $\Lambda$ from per-document squared LoRA gradients on the reference documents; (iii)~the perplexity gate, a forward pass on every scanned candidate; (iv)~per-sample LoRA gradients, computed only for candidates that pass the gate; (v)~the TRAK projection of the anchor and frontier signatures to $d=8192$ dimensions; (vi)~Sieve-Streaming selection with Sherman--Morrison updates; and (vii)~CPT on the selected tokens, with the same LoRA recipe as every baseline.

\begin{table}[t]
\centering
\caption{\textbf{End-to-end compute of our pipeline} on TinyLlama-1.1B at the default $4$B-token budget, measured on $4$ nodes with $4\times$ A100-40G GPUs each. GPU-hours are summed over all GPUs; the share is the fraction of total GPU-hours.}
\label{tab:compute-stages}
\small
\begin{tabular}{@{}l r r r@{}}
\toprule
Stage & EFLOP & GPU-h & Share \\
\midrule
(i) Warmup adapter $\theta^\star$ & $\phantom{0}1.30$ & $5.9$ & $3.2\%$ \\
(ii) Diagonal Fisher $\Lambda$ & $\phantom{0}1.27$ & $24.7$ & $13.4\%$ \\
(iii) Perplexity gate & $\phantom{0}3.06$ & $13.7$ & $7.4\%$ \\
(iv) Per-sample LoRA gradients & $\phantom{0}3.36$ & $26.7$ & $14.5\%$ \\
(v) TRAK projection & $\phantom{0}1.88$ & $4.4$ & $2.4\%$ \\
(vi) Sieve-Streaming selection & $<0.01$ & $1.1$ & $0.6\%$ \\
(vii) CPT & $24.6\phantom{0}$ & $107.5$ & $58.4\%$ \\
\midrule
\textbf{Total} & $\mathbf{35.5}\phantom{0}$ & $\mathbf{184.0}$ & \\
\bottomrule
\end{tabular}
\end{table}

\noindent\textbf{Where the cost goes.}
Training, not selection, is the largest single cost: CPT accounts for $58.4\%$ of the GPU-hours at the $4$B budget, and stages (i)--(vi) together for the remaining $76.5$ GPU-hours. Within selection, the two per-document gradient stages dominate: the Fisher estimate (ii) and the per-sample candidate gradients (iv) take $24.7$ and $26.7$ GPU-hours. They also run at much lower hardware utilization than CPT: stage (ii) performs $1.27$ EFLOP in $24.7$ GPU-hours, whereas CPT performs $24.6$ EFLOP in $107.5$ GPU-hours, likely because these stages run unpacked and compute one backward pass per document (App.~\ref{sec:speedups:stream}). The Sieve-Streaming step itself, the only stage whose cost is measured in Table~\ref{tab:streaming-approx}, is negligible ($0.6\%$), and the perplexity gate keeps the backward pass of stage (iv) off the rejected candidates.

\begin{table}[t]
\centering
\caption{\textbf{End-to-end cost against Replay} (TinyLlama-1.1B). Totals include every selection stage of Table~\ref{tab:compute-stages}; Replay performs no selection. The ratio columns give Replay-$10$B's cost divided by the method's cost. $A$ and $\Phi$ are from Tables~\ref{tab:main} and~\ref{tab:token-efficiency}.}
\label{tab:compute-replay}
\small
\setlength{\tabcolsep}{5pt}
\begin{tabular}{@{}l c r r c c c c@{}}
\toprule
Method & CPT tokens & EFLOP & GPU-h & FLOP ratio & GPU-h ratio & $A\uparrow$ & $\Phi\downarrow$ \\
\midrule
Replay & $10$B & $62.1$ & $271.4$ & $1.0\times$ & $1.0\times$ & $-$2.0 & 0.8 \\
Ours   & $4$B  & $35.5$ & $184.0$ & $1.7\times$ & $1.5\times$ & +5.5 & 0.4 \\
Ours   & $1$B  & $10.9$ & $69.3$  & $5.7\times$ & $3.9\times$ & +3.0 & 0.6 \\
\bottomrule
\end{tabular}
\end{table}

\noindent\textbf{Comparison with Replay.}
Table~\ref{tab:compute-replay} compares the full cost of our pipeline with Replay at $10$B tokens, which spends $62.1$ EFLOP and $271.4$ GPU-hours. At the $1$B budget of the token-efficiency comparison, our pipeline costs $10.9$ EFLOP and $69.3$ GPU-hours end to end, including selection. It thus uses $5.7\times$ fewer FLOPs and $3.9\times$ fewer GPU-hours than Replay at $10$B, while outperforming it on both adaptation ($A=+3.0$ vs.\ $-2.0$) and forgetting ($\Phi=0.6$ vs.\ $0.8$). At the default $4$B budget, the pipeline remains cheaper than Replay at $10$B ($1.7\times$ fewer FLOPs and $1.5\times$ fewer GPU-hours) and improves $A$ by $7.5$ points. The GPU-hour ratios are smaller than the FLOP ratios because the per-document gradient stages (ii) and (iv) run at lower utilization than CPT. The token-efficiency advantage of \S\ref{sec:exp:analysis} therefore carries over to end-to-end compute, but by a smaller factor ($3.9\times$ in GPU-hours at $1$B) than the $10\times$ ratio of training tokens.

\subsection{Additional ablations}

\paragraph{Fisher Floor.}
\label{app:abl-weights-floor:floor}
The frontier view divides by $\sqrt{\Lambda_\varepsilon}$ with $\Lambda_{\varepsilon,i}=\max\{\Lambda_i,\varepsilon_\Lambda\}$, so coordinate $i$ is amplified by $\Lambda_{\varepsilon,i}^{-1/2}$. This is where the frontier branch is numerically most fragile. If the floor is too low, the smallest entries of $\Lambda$, which are also the least reliably estimated, receive the largest weights, and a few amplified coordinates can dominate $\|\gfr(x)\|_2$ and the frontier scatter matrix. The floor caps the amplification at $\varepsilon_\Lambda^{-1/2}$. If the floor is too high, every clamped coordinate receives the same weight, which erases the contrast between low- and very-low-Fisher directions; in the limit $\varepsilon_\Lambda\ge\max_i\Lambda_i$, $\gfr(x)=\hat g(x)/\sqrt{\varepsilon_\Lambda}$ and the frontier term reduces to an identity-geometry $\logdet$ on the raw preconditioned gradient. The floor does not enter the anchor view $\ganc=\sqrt{\Lambda}\odot\hat g$.

Table~\ref{tab:abl-floor} sweeps the floor over percentiles $p_q(\Lambda)$ of the Fisher entries, together with the fixed floor $10^{-8}$, which clips no coordinate and is therefore no effective floor. Without an effective floor, the frontier norm is inflated by up to $3.1\times10^4$ and the method degrades to $A=+3.0$, $\Phi=4.5$, consistent with a few amplified coordinates dominating the frontier term. A $5$th-percentile floor already removes much of this damage ($\Phi$ drops from $4.5$ to $1.9$), and every floor between the $25$th and $75$th percentiles stays within $0.3$ of the default in $A$ and within $0.1$ in $\Phi$, so the method is not sensitive to the exact choice of the median. At the other extreme, a constant $\Lambda$ ($p_{100}$) removes the Fisher reweighting from the frontier view and drops $A$ to $+3.4$ while raising $\Phi$ to $1.2$. This row also rescales the whole frontier signature by $(\max_i\Lambda_i)^{-1/2}$, which by the rescaling equivalence above changes the effective $\gamma$, so it mixes the loss of Fisher contrast with a change of scale; it nevertheless suggests that the $\Lambda^{-1/2}$ reweighting itself, not only the clipping, contributes to the gain.

\begin{table}[t]
\centering
\caption{\textbf{Fisher floor $\varepsilon_\Lambda$} (TinyLlama-1.1B, $4$B tokens). $p_q(\Lambda)$ is the $q$-th percentile of the Fisher entries; ``coords clipped'' is the fraction of coordinates raised to the floor; the inflation column reports the largest inflation of $\|\gfr\|$ relative to the constant-$\Lambda$ row. $^\star$ marks the default.}
\label{tab:abl-floor}
\small
\setlength{\tabcolsep}{5pt}
\begin{tabular}{@{}l c c c c@{}}
\toprule
$\varepsilon_\Lambda$ & Coords clipped & Max $\|\gfr\|$ inflation & $A\uparrow$ & $\Phi\downarrow$ \\
\midrule
$10^{-8}$ (no effective floor)                 & $0.0\%$   & $3.1\times10^{4}\times$ & +3.0 & 4.5 \\
$p_{5}(\Lambda)$                               & $5\%$     & $210\times$             & +4.1 & 1.9 \\
$p_{25}(\Lambda)$                              & $25\%$    & $21\times$              & +5.3 & 0.5 \\
$p_{50}(\Lambda)$ = median$^\star$             & $50\%$    & $6.4\times$             & +5.5 & 0.4 \\
$p_{75}(\Lambda)$                              & $75\%$    & $2.3\times$             & +5.2 & 0.4 \\
$p_{100}$ (constant $\Lambda$; frontier on $\hat g$) & $100\%$ & $1.0\times$           & +3.4 & 1.2 \\
\bottomrule
\end{tabular}
\end{table}

\paragraph{Hyperparameter sensitivity.}
Table~\ref{tab:sensitivity} shows that the default hyperparameters are robust and near-optimal across all four sweeps. The default setting attains the strongest adaptation gain while keeping forgetting low, and nearby settings preserve most of the gain. Performance degrades mainly under deliberately weak configurations: removing warmup ($T=0$) collapses adaptation, using too small a projection dimension increases forgetting, and small per-term scales reduce the benefit of the $\logdet$ objective.

\begin{table}[t]
\centering
\caption{Hyperparameter sensitivity on TinyLlama-1.1B under the 4B-token default setting. Each block sweeps one hyperparameter holding the rest at their defaults (marked $^\star$). Last column reports the best TinyLlama baseline from Table~\ref{tab:main} under the same $A-\Phi$ metric.}
\label{tab:sensitivity}
\small
\setlength{\tabcolsep}{5pt}
\begin{tabular}{l c c c}
\toprule
Sweep & Value & $A\uparrow$ & $\Phi\downarrow$ \\
\midrule
\multirow{4}{*}{Frontier weight $\alpha$}
  & 0.5            & +4.8 & 0.7 \\
  & 1.0$^\star$    & \textbf{+5.5} & \textbf{0.4} \\
  & 2.0            & +5.4 & 0.5 \\
  & 4.0            & +4.6 & 0.4 \\
\midrule
\multirow{4}{*}{Per-term scale $\beta=\gamma$}
  & 5              & +3.8 & 1.0 \\
  & 10             & +4.9 & 0.6 \\
  & 20$^\star$     & \textbf{+5.5} & 0.4 \\
  & 50             & +5.4 & 0.4 \\
\midrule
\multirow{4}{*}{Warmup steps $T$}
  & 0              & +0.6 & 0.5 \\
  & 100            & +4.7 & 0.5 \\
  & 200$^\star$    & \textbf{+5.5} & 0.4 \\
  & 500            & +5.5 & 0.4 \\
\midrule
\multirow{4}{*}{TRAK dim $d$}
  & 1024           & +4.0 & 1.2 \\
  & 4096           & +5.3 & 0.5 \\
  & 8192$^\star$   & \textbf{+5.5} & 0.4 \\
  & 16384          & +5.5 & 0.4 \\
\bottomrule
\end{tabular}
\end{table}

\paragraph{Streaming approximation quality and per-sample cost.}

\begin{table}[t]
\centering
\caption{Small-stream Sieve-Streaming diagnostic on cached projected gradients ($|\Stream| = 10^4$, $B = 1{,}000$, TinyLlama-1.1B). Offline greedy is used only to normalize the projected objective in this tractable regime; no downstream CPT is run for this diagnostic. $|\mathcal T|$ is the number of sieve thresholds.}
\label{tab:streaming-approx}
\small
\begin{tabular}{c c c c}
\toprule
$\varepsilon$ & $|\mathcal T|$ & $f(S_{\text{ours}})/f(S_{\text{greedy}})$ & per-candidate wall-clock (ms) \\
\midrule
0.05 & 94 & 99.1\% & 7.8\phantom{00} \\
0.1  & 48 & 98.5\% & 4.0\phantom{00} \\
0.2  & 25 & 96.8\% & 2.1\phantom{00} \\
0.4  & 14 & 94.6\% & 1.2\phantom{00} \\
\midrule
\multicolumn{2}{l}{Offline greedy (diagnostic reference)} & 100\% & 720\phantom{.0} \\
\bottomrule
\end{tabular}\vspace{-0.6cm}
\end{table}
\section{Limitations}
\label{sec:limitations}

Our method is designed for scalable CPT data selection, so its approximations are deliberately chosen to preserve the main Fisher geometry while remaining deployable for billion-parameter LMs. In particular, we estimate curvature with a diagonal Fisher in the LoRA adapter subspace, evaluate gradients at a short warmup operating point $\theta^\star$, and preserve the subset-level log-determinant geometry through random projection rather than full-parameter covariance matrices. These choices are part of the scalable formulation and are examined through the ablations and streaming diagnostics in \S\ref{sec:exp:ablation} and \S\ref{sec:exp:analysis}; they also leave natural extensions to richer block-diagonal or low-rank curvature models, full-parameter adaptation, and adaptive warmup schedules. Our experiments focus on medical-domain CPT for two LLM backbones, so the empirical claims are scoped to this setting, with other domains, instruction-tuning regimes, and multimodal streams left for future validation.

\section{Extended Related Work}
\label{app:related}

This appendix expands the compressed discussion of \S\ref{sec:related} with the descriptions and method-level distinctions that were elided for space. All citations referenced here also appear in the main related work.

\noindent\textbf{Continual pre-training of large language models.}
Continual pre-training (CPT) adapts a pretrained LLM to a new domain, language, or temporal slice without retraining from scratch, and the dominant concern is catastrophic forgetting of capabilities acquired during the original run. A first line of work stabilizes the optimizer itself: a combination of learning-rate re-warming, re-decay, and replay of a small fraction of prior data is sufficient to handle weak (English$\to$English) and strong (English$\to$German) distribution shifts \citep{ibrahim2024simple}, and follow-up work further layers topic-based mixture reweighting and a perplexity-based curriculum on top \citep{chen2025towards}. Domain-specialized CPT has been studied in finance \citep{xie2024efficient} and in mathematics and coding \citep{fujii2025rewriting}, and scaling-law analyses of the domain/general trade-off are given by the D-CPT law \citep{que2024d}.

Another thread explicitly analyzes forgetting during CPT of aligned or instruction-tuned models \citep{li2024examining,yildiz2024investigating,shi2025continual}, and several methods mitigate it by routing or model-merging rather than by data: distribution-specialized experts \citep{chen2023lifelong}, language-inclusive CLIP fine-tuning \citep{yang2024embracing}, task arithmetic for selective forgetting \citep{ilharco2022editing}, WiSE-FT-style robust fine-tuning \citep{wortsman2022robust}, and the Soup-to-Go model-averaging recipe \citep{kleiman2025soup}. CPT has also been extended to knowledge-intensive generative models \citep{guo2025corpusbrain}, to multimodal LLMs \citep{zhao2025mllm,roth2024practitioner}, and to architecture-level adaptation from autoregressive to diffusion language models \citep{gong2024scaling}.

Our contribution inverts the role of the Fisher relative to EWC. EWC \citep{kirkpatrick2017overcoming} uses the diagonal Fisher as a \emph{regularizer} that pins parameters near their pretrained values after the data is fixed. We use the same Fisher upstream of optimization, to decide which tokens enter the CPT stream, so adaptation can proceed with an unconstrained optimizer and no auxiliary penalty.

\noindent\textbf{Data selection for language-model training.}
Selecting training data for LMs has been investigated at every training stage. For from-scratch pre-training, DoReMi optimizes domain mixtures via group DRO on a proxy model \citep{xie2023doremi}; DSIR reframes the choice as importance resampling against a target distribution \citep{xie2023data}; and a broad family of scorers learns to rank candidates, including data-influence models (MATES \citep{yu2024mates}), group-level selection \citep{yu2025group}, quality classifiers (DataMan \citep{peng2025dataman}), meta-learned raters (DataRater \citep{calian2025datarater}), differentiable mixture weights \citep{tanfast}, relative scoring via Bradley--Terry comparisons \citep{chen2025murating}, sensitivity-aware greedy selection (GREATS \citep{wang2024greats}), autonomous LM-based selection for mathematical text \citep{zhang2024autonomous}, and large-scale data pruning at pre-training scale \citep{marion2023less,longpre2024pretrainer,du2022glam}.

A parallel thread targets \emph{instruction tuning}, where strong results are obtained with dramatically smaller curated sets: LIMA and its ``0.5\% is enough'' successor \citep{zhou2023lima,chen2023maybe}, AI-scored filtering in AlpaGasus \citep{chen2023alpagasus}, alignment-quality studies \citep{liu2023makes}, LIFT \citep{xu2023rethinking}, unsupervised core-set selection DEFT \citep{das2023deft}, code-oriented pruning \citep{tsai2024code}, visual instruction-tuning selectors \citep{liu2025less,wu2024icons,wang2025data}, influence distillation \citep{nikdan2025efficient}, active and self-guided schemes \citep{kung2023active,li2024quantity}, task-aware distribution matching \citep{chengtask}, activation-stability pruning \citep{alambatch}, difficulty--diversity collaborative filtering \citep{hoangdifficulty}, tree-structured synthesis \citep{wang2025treesynth}, and a growing list of IT-focused selectors \citep{li2024superfiltering}. More broadly, loss- and difficulty-based pruning has a long pedigree in classification and NLU \citep{sorscher2022beyond,paul2021deep,attendu2023nlu,nguyen2025synthetic}.

Our selector differs from these methods on two axes. First, it operates in the post-pretraining regime where there is prior competence to protect, unlike pre-training selectors that assume a \emph{tabula rasa}, and outside the small-budget instruction-tuning regime targeted by LIMA-style filters. Second, it decomposes a sample's expected contribution into an anchor component aligned with high-Fisher directions of established competence and a frontier component aligned with low-Fisher directions where the pretrained loss is flat. This curvature-weighted decomposition is invisible to loss, perplexity, and feature-similarity surrogates, which integrate over directions without weighting them by Fisher mass.

\noindent\textbf{Coreset selection and submodular data subset methods.}
Our global selection formulation builds on the coreset literature surveyed in DeepCore \citep{guo2022deepcore}. Geometry-based methods retain representative points via herding \citep{welling2009herding} or $k$-center-greedy coverage \citep{farahani2009facility,sener2017active}; uncertainty- and loss-based methods use forgetting events \citep{toneva2018empirical}, gradient norms and error norms (GraNd, EL2N) \citep{paul2021deep}, or sensitivity-based importance sampling; decision-boundary methods exploit adversarial deep-fool margins and contrastive active learning \citep{ducoffe2018adversarial,margatina2021active}; gradient-matching methods greedily select subsets whose aggregate gradient tracks the full training gradient (CRAIG \citep{mirzasoleiman2020coresets}, GRAD-MATCH \citep{killamsetty2021grad}) and give OMP-based submodular approximation guarantees; and bilevel methods treat the subset as an inner optimization variable (GLISTER \citep{killamsetty2021glister}, RETRIEVE \citep{killamsetty2021retrieve}). Data pruning at neural-scaling-law scale \citep{sorscher2022beyond} established that well-chosen subsets can beat power-law scaling, and the dataset-distillation line \citep{wang2018dataset,zhao2020dataset} condenses rather than selects. We also draw inspiration from simple local-rule emergent coordination \citep{reynolds1987flocks}.

Our setting and utility differ from these methodd, where the per-candidate utility is derived from the pretrained model's Fisher curvature rather than from gradient magnitude, feature geometry, or loss trajectories, and the resulting constant-factor approximation has per-sample cost independent of the candidate-pool size.

\end{document}